\documentclass[sn-mathphys-num]{sn-jnl}

\usepackage{graphicx}%
\usepackage{multirow}%
\usepackage{amsmath,amssymb,amsfonts}%
\usepackage{amsthm}%
\usepackage{mathrsfs}%
\usepackage[title]{appendix}%
\usepackage{xcolor}%
\usepackage{textcomp}%
\usepackage{manyfoot}%
\usepackage{booktabs}%
\usepackage{algorithm}%
\usepackage{algorithmicx}%
\usepackage{algpseudocode}%
\usepackage{listings}%

\theoremstyle{thmstyleone}%
\theoremstyle{thmstyletwo}%

\theoremstyle{thmstylethree}%

\begin{document}

\title[Article Title]{Interpreting the Curse of Dimensionality from Distance Concentration and Manifold Effect}


\author[1,2,3]{\fnm{Dehua} \sur{Peng}}\email{pengdh@whu.edu.cn}

\author*[1,3]{\fnm{Zhipeng} \sur{Gui}}\email{zhipeng.gui@whu.edu.cn}

\author[2,3]{\fnm{Huayi} \sur{Wu}}\email{wuhuayi@whu.edu.cn}

\affil*[1]{\orgdiv{School of Remote Sensing and Information Engineering}, \orgname{Wuhan University}, \orgaddress{\city{Wuhan}, \postcode{430079}, \state{Hubei}, \country{China}}}

\affil[2]{\orgdiv{State Key Laboratory of Information Engineering in Surveying, Mapping and Remote Sensing}, \orgname{Wuhan University}, \orgaddress{\city{Wuhan}, \postcode{430079}, \state{Hubei}, \country{China}}}

\affil[3]{\orgdiv{Collaborative Innovation Center of Geospatial Technology}, \orgname{Wuhan University}, \orgaddress{\city{Wuhan}, \postcode{430079}, \state{Hubei}, \country{China}}}

\abstract{The characteristics of data like distribution and heterogeneity, become more complex and counterintuitive as dimensionality increases. This phenomenon is known as curse of dimensionality, where common patterns and relationships (e.g., internal pattern and boundary pattern) that hold in low-dimensional space may be invalid in higher-dimensional space. It leads to a decreasing performance for the regression, classification, or clustering models or algorithms. Curse of dimensionality can be attributed to many causes. In this paper, we first summarize the potential challenges associated with manipulating high-dimensional data, and explains the possible causes for the failure of regression, classification, or clustering tasks. Subsequently, we delve into two major causes of the curse of dimensionality, distance concentration, and manifold effect, by performing theoretical and empirical analyses. The results demonstrate that, as the dimensionality increases, nearest neighbor search (NNS) using three classical distance measurements, Minkowski distance, Chebyshev distance, and cosine distance, becomes meaningless. Meanwhile, the data incorporates more redundant features, and the variance contribution of principal component analysis (PCA) is skewed towards a few dimensions.}




\maketitle

\section{Introduction}
\label{sec1}
With the rapid development of data collection and storage technologies, we have entered the era of big data, where the data holds a trend of rapid growth in both sample size and feature dimensionality \cite{ref1}, \cite{ref2}, \cite{ref3}. However, directly dealing with the data in high-dimensional feature space faces the curse of dimensionality \cite{ref4}, \cite{ref5}, including the following challenges.

\begin{figure}[h]
\centering
\includegraphics[width=0.95\linewidth]{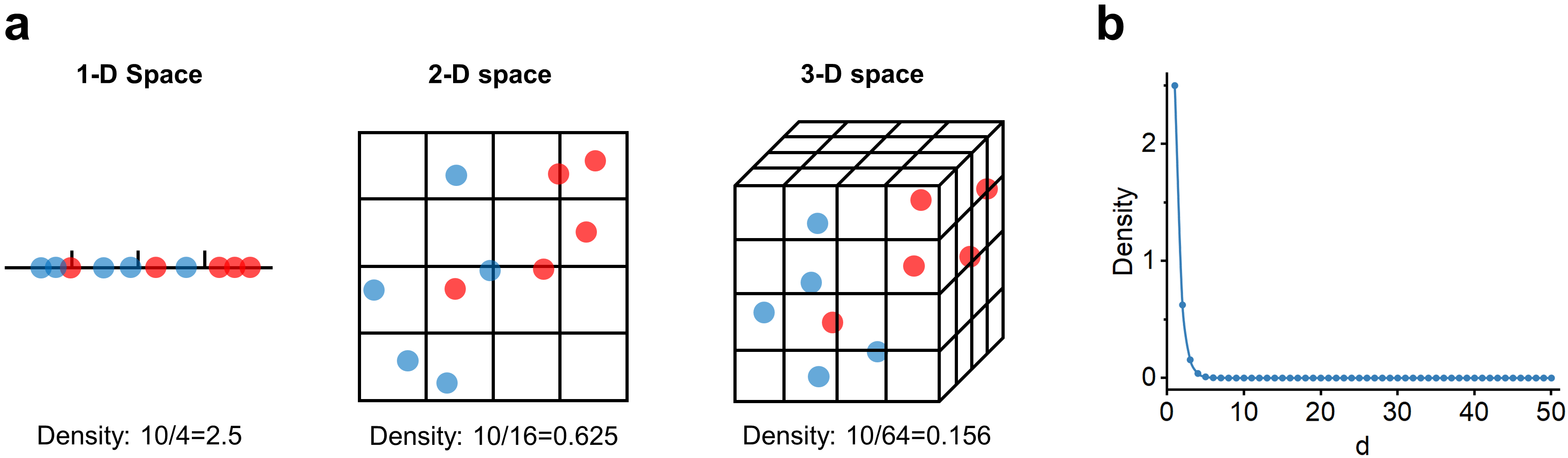}
\caption{An example of ten data samples for illustrating the data sparsity in high-dimensional space. (a) Data distributions of the samples in 1-D to 3-D feature space. (b) The trend of sample density as the dimension increases.}
\label{fig1}
\end{figure}

The distribution of data samples in high-dimensional feature space commonly exhibits severe sparsity, which leads to the model’s inability to represent the entire feature space \cite{ref4}, \cite{ref6}, \cite{ref7}. To explain this phenomenon, let us consider a simple example. Ten samples are given in Fig. \ref{fig1}\textcolor{blue}{a}, and each data feature is divided into four intervals. We can calculate the sample density for each feature interval, which is 2.5 in 1-D space, while the density decreases to 0.625 and 0.156 in 2- and 3-D spaces, respectively. We can infer that the sample density is $10/4^d$ in a $d$-dimensional space. As $d$ approaches infinity, the density converges to 0. From Fig. \ref{fig1}\textcolor{blue}{b}, we can observe that when $d$ is larger than 5, the density is already close to 0. Typically, the number of samples $n$ is significantly smaller than $4^d$. It implies that most of the feature intervals do not contain any samples. This makes it hard for the models to comprehensively learn and represent the feature space.

Data sparsity in high-dimensional space causes the models overfitting and weakens the generalization performance \cite{ref8}, \cite{ref9}. To classify data samples, a classifier needs to trained on some annotated samples for learning and representing the features. Then, the trained model is applied to non-annotated samples for validation. Taking the support vector machine (SVM) classifier as an example, it generates an optimal decision hyperplane in the training samples \cite{ref10}. However, if the samples are too sparse, although this nonlinear decision surface can obtain a high training accuracy, the model performance will be significantly compromised when applied to non-training data samples, especially those with significant differences in attributes compared to the training samples, due to insufficient representation ability. One way to address overfitting problem is to increase the number of data samples. However, in practical applications, the available data samples are often limited. 

Distance measurement may be invalid in high-dimensional space due to the phenomenon of distance concentration \cite{ref5}, \cite{ref11}, \cite{ref12}, \cite{ref13}. It refers to that the pair-wise distances between different data points converge to the same value as the dimensionality increases \cite{ref14}, \cite{ref15}, \cite{ref16}. Commonly, machine learning tasks adopt distance proximity to measure the similarity between data samples. For example, density-based spatial clustering of applications with noise (DBSCAN) identifies the clusters by connecting the high-density circular units with a fixed neighborhood distance \cite{ref17}. Distance concentration makes the optimal Eps of DBSCAN more difficult to specify in a higher-dimensional space. Another example is the K-nearest neighbors (KNN) classifier, which considers that each point has the highly similar label to its KNN \cite{ref18}. Distance concentration makes the similarity between different points become ambiguous, which severely affects the effectiveness of KNN classifier.

\begin{figure}[h]
\centering
\includegraphics[width=0.95\linewidth]{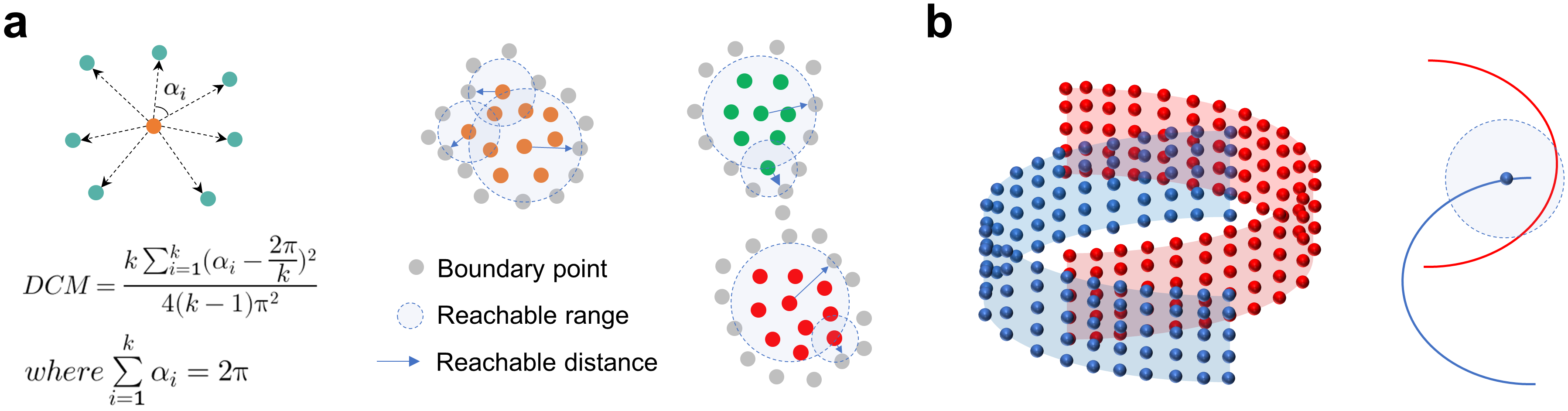}
\caption{Manifold structure has an undesirable effect on boundary-seeking clustering algorithms. (a) Illustration of the CDC algorithm. (b) Boundary-based constraint cannot prevent cross-cluster connections in manifolds.}
\label{fig2}
\end{figure}

The manifold structure of high-dimensional data is not conducive to classification and clustering tasks. High-dimensional data often contains a bunch of nonlinear manifolds, which have no distinguishable gaps and are hard to be separated \cite{ref19}, \cite{ref20}. Such a manifold effect introduces difficulties on two fronts. On the one hand, a non-linear manifold structure like a Swiss Roll cannot be represented by Euclidean-based similarity \cite{ref21}, \cite{ref22}. For instance, spectral clustering produces skew graph cuts in manifold-shaped clusters using Euclidean-based distances \cite{ref23}. Path-based similarity is probably more suitable for capturing the topological structure of manifolds and ensuring the strong associations between the points in the same manifold. On the other hand, manifolds do not have the concepts of boundary and internal points \cite{ref24}. Thus, the algorithms that detect the structure of clusters by identifying the boundary and internal patterns become invalid. For example, clustering using direction centrality (CDC) algorithm determines the boundary points through angle variance, and then generates clusters by connecting the internal points \cite{ref25} (Fig. \ref{fig2}\textcolor{blue}{a}). Its core idea lies in that boundary points can generate enclosed cages to bind the connections of internal points, thereby preventing cross-cluster connections and separating weakly-connected clusters. However, the identified boundary points of manifold structured clusters fail to form an all-direction constraint for the internal connections, which makes adjacent clusters cannot be separated (Fig. \ref{fig2}\textcolor{blue}{b}).

Excessive redundant features impose burdens on data storage and computation. Users often intend to expand the feature dimensionality for higher precision, however, such operations would introduce excessive redundant features. These features are either highly linearly correlated, or present the same values across all samples, or are noisy that interfere with classification and clustering \cite{ref8}. Meanwhile, excessive features increase the time and space complexity of algorithms, thereby placing higher demands on computational resources and calling for more efficient algorithms. We take KNN search as an example. The brute force method can be divided into two steps, pair-wise distance computation and selecting the $K$ smallest distances. It should consider the number of dimensions $d$ to computing the pair-wise distances, since more directions mean longer vectors to compute distances \cite{ref26}. Obviously, the time efficiency of KNN search is related to data dimensionality $d$. Higher data dimensions, lower computational efficiency.

In this paper, we interpret the curse of dimensionality from the perspective of distance concentration and manifold effect through theoretical and empirical analyses The major contributions of our work can be summarized as follows:

\begin{itemize} 
  \item We expand the theoretical proof of the concentration effect of Minkowski distance and reveal that the Chebyshev distance and cosine distance measurements also exhibit a concentration phenomenon.
  \item We discover that high-dimensional data inevitably presents a manifold structure through the asymptotic behavior of PCA.
  \item In addition to theoretical analysis, we also validate the findings of this paper through simulation experiments and real-world datasets.
  
\end{itemize}

This paper is organized as follows: Section \ref{sec2} provide necessary mathematics. Section \ref{sec3} and Section \ref{sec4} theoretically prove the distance concentration and manifold effect, respectively. While Section \ref{sec5} draws the conclusion.

\section{Background Mathematics}
\label{sec2}
First of all, we would like to remind necessary background mathematical concepts and lemmas in this section.

\textbf{Central Limit Theorem:} Suppose that $x_1,x_2,…x_n$ are independent and identically distributed (IID) random variables with mean $\mu$ and variance $\sigma^{2}$, when $n$ is large, we have
\begin{equation}
\left. {\sum_{i = 1}^{n}x_{i}} \right.\sim\mathcal{N}\left( {n\mu,~n\sigma^{2}} \right)
\label{eq1}
\end{equation}

\textbf{Slutsky’s Theorem:} Given a sequence of random variables $x_1,x_2,…x_n$ and a continuous function $G$. If $x_{n}\overset{p}{\rightarrow}c$ and $G(c)$ is finite then $G\left( x_{n} \right)\overset{p}{\rightarrow}G(c)$.

\textbf{Limit of a Sequence:} Given a sequence $\left\{ x_{n} \right\}$ and a real constant $c \in \mathbb{R}$, if there exists a positive integer $N \in Z^{+}$, for every $\varepsilon > 0$ such that $\left| {x_{n} - c} \right| \leq \varepsilon$ for every $n>N$, we say that the sequence $\left\{ x_{n} \right\}$ converges $c$, which can be written as 
\begin{equation}
    {\lim\limits_{n\rightarrow\infty}x_{n}} = c
\label{eq2}
\end{equation}

\textbf{Cauchy Interlace Theorem:} Let $\mathbf{X} \in \mathbb{R}^{d \times d}$ be a Hermitian matrix of order $d$, and let $\mathbf{Y}$ be a principal submatrix of $\mathbf{X}$ of order $d-1$. If the eigenvalues of $\mathbf{X}$ are arranged as $\lambda_{1}\left( \mathbf{X} \right) \leq \lambda_{2}\left( \mathbf{X} \right) \leq \ldots \leq \lambda_{d}\left( \mathbf{X} \right)$, and $\lambda_{1}\left( \mathbf{Y} \right) \leq \lambda_{2}\left( \mathbf{Y} \right) \leq \ldots \leq \lambda_{d - 1}\left( \mathbf{Y} \right)$ are the eigenvalues of $\mathbf{Y}$, then we have $\lambda_{i}\left( \mathbf{X} \right) \leq \lambda_{i}\left( \mathbf{Y} \right) \leq \lambda_{i + 1}\left( \mathbf{X} \right)$, for $i=1,2,…d-1$.

\textbf{Lemma 1:} Given a sequence of random variables with finite variance $\mathbf{X} = \left\{ x_{1},x_{2},\ldots x_{n} \right\}$, if ${\lim_{n\rightarrow\infty}{\mathbf{E}\left( \mathbf{X} \right)}} = c$ and ${\lim_{n\rightarrow\infty}{var\left( \mathbf{X} \right)}} = 0$, then we will have ${\lim_{n\rightarrow\infty}x_{n}} = c$. 

\textbf{\textit{Proof:}} Suppose that as $n$ approaches positive infinity, $x_n$ does not have a limit and does not converge to the constant $c$. Thus, there exists a positive real $\varepsilon > 0$, for every $N \in Z^{+}$, there exists a positive integer $N \in Z^{+}$ such that $\left| {x_{n} - c} \right| > \varepsilon$.

Based on the definition of the sequence limit, given a positive real $0.5\varepsilon > 0$, there exists $N_{1} \in Z^{+}$ such that $\left| {\mathbf{E}\left( \mathbf{X} \right) - c} \right| \leq 0.5\varepsilon$ for every $n>N_1$. According to the corollary, there exists a positive integer $n_1>N_1$ that makes
\begin{equation}
    \left| {x_{n_{1}} - c} \right| > \varepsilon
\label{eq3}
\end{equation}

So, we have 
\begin{equation}
    \frac{1}{n_{1}}\left( {x_{n_{1}} - \mathbf{E}\left( \mathbf{X} \right)} \right)^{2} > \frac{\varepsilon^{2}}{{4n}_{1}}
\label{eq4}
\end{equation}

Besides, there exists $N_{2} \in Z^{+}$, for every $n>N_2$, it holds that
\begin{equation}
    var\left( \mathbf{X} \right) \leq \frac{\varepsilon^{2}}{{4n}_{1}}
\label{eq5}
\end{equation}

$i$) If $n_1>N_2$, we let $n=n_1$ and then have
\begin{equation}
\begin{aligned}
    var\left( \mathbf{X} \right) = \frac{1}{n_{1}}{\sum\limits_{i = 1}^{n_{1}}\left( {x_{i} - \mathbf{E}\left( \mathbf{X} \right)} \right)^{2}} \geq \frac{1}{n_{1}}\left( {x_{n_{1}} - \mathbf{E}\left( \mathbf{X} \right)} \right)^{2} > \frac{\varepsilon^{2}}{{4n}_{1}}
\end{aligned}
\label{eq6}
\end{equation}

Eq. \eqref{eq5} and \eqref{eq6} are in conflict, so the assumption does not hold.

$ii$) If $N_{2} \geq n_{1} > N_{1}$, so there exists $n_2>N_2$ such that
\begin{equation}
    \left| {x_{n_{2}} - c} \right| > \varepsilon,~~\left| {\mathbf{E}\left( \mathbf{X} \right) - c} \right| \leq 0.5\varepsilon
\label{eq7}
\end{equation}

\begin{figure}[h]
\centering
\includegraphics[width=0.95\linewidth]{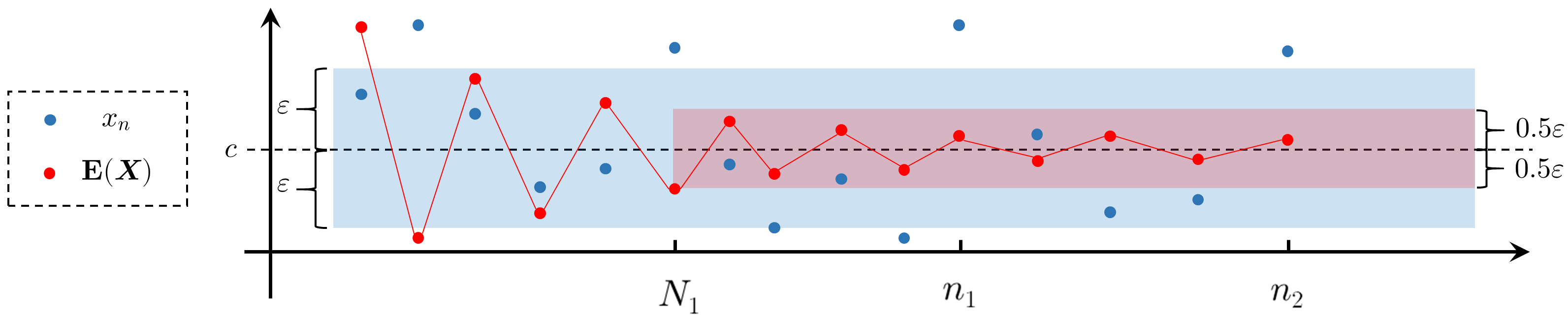}
\caption{Illustration for proving Lemma 1.}
\label{fig3}
\end{figure}

Similarly, there also exists a contradiction as deduced in Eq. \eqref{eq4}-\eqref{eq6}, and an illustration is shown in Fig. \ref{fig3}.

\section{Distance Concentration}
The concentration law of Minkowski distance has been investigated and proved in the previous works \cite{ref11}, \cite{ref12}. In this section, we optimize the proof of Minkowski distance and extend this law to more typical distance measurements, such as Chebyshev distance and cosine distance.
\label{sec3}
\subsection{Minkowski Distance}

$L_k$-norm Minkowski distance is a widely-used dissimilarity measurement between different data points and is a generalization of the Manhattan distance ($k=1$), Euclidean distance ($k=2$), and Chebyshev distance ($k = \infty$). To explain the distance concentration, we take an intuitive use case, NNS, inspired by the research works in \cite{ref11}, \cite{ref12}. When the dimensionality $d$ is low, there is commonly a significant difference in distances, NNS is meaningful. However, with the increase of $d$, the distances of different point pairs converge to the same value, and the discrimination decreases. NNS becomes meaningless. The distance concentration phenomenon in Minkowski distance can be formally depicted as the following theorem:

\textbf{Definition:} Given $n$ data points $P_d^1,P_d^2,…P_d^n$, where $P_{d}^{i} = \left( p_{1}^{i},p_{2}^{i},\ldots p_{d}^{i} \right) \in \mathbb{R}^{d}$, and a query point $Q_{d} = \left( q_{1},q_{2},\ldots q_{d} \right) \in \mathbb{R}^{d}$, we can calculate the $L_k$-norm Minkowski distance between the data point $P_d^i$ and the query point $Q_d$ as
\begin{equation}
    \left\| {P_{d}^{i} - Q_{d}} \right\|_{k} = \left( {\sum_{j = 1}^{d}\left| {p_{j}^{i} - q_{j}} \right|^{k}} \right)^{\frac{1}{k}}
\label{eq8}
\end{equation}

We define 
\begin{equation}
\begin{aligned}
    D_{d,min} = {\min\left\{ {\left\| {P_{d}^{i} - Q_{d}} \right\|_{k}~} \middle| {i = 1,2,\ldots,n} \right\}}
    \\D_{d,max} = {\max\left\{ {\left\| {P_{d}^{i} - Q_{d}} \right\|_{k}~} \middle| {i = 1,2,\ldots,n} \right\}}
\end{aligned}
\label{eq9}
\end{equation}

\textbf{Theorem 1:} If each dimension of $P_d$ and $Q_d$ is IID, we have the relative distance ratio (RDR)
\begin{equation}
    \lim\limits_{d\rightarrow\infty}RDR = \lim\limits_{d\rightarrow\infty}\frac{\left| {D_{d,max} - D_{d,~min}} \right|}{D_{d,~min}} = 0
\label{eq10}
\end{equation}

\textbf{\textit{Proof:}} For the sake of simplicity, we prove Theorem 1 using a specific example. If there exists two data points $A_{d} = \left( a_{1},a_{2},\ldots a_{d} \right) \in \mathbb{R}^{d}$ and $B_{d} = \left( b_{1},b_{2},\ldots b_{d} \right) \in \mathbb{R}^{d}$, where $a_{i},b_{i} \sim U(0,~1)$, and the query point is $Q_{d} = (0,0,\ldots 0) \in \mathbb{R}^{d}$, then we can compute the $L_k$-norm Minkowski distances to the query point as
\begin{equation}
\begin{aligned}
    A_{d}Q_{d} = \left( {\sum_{i = 1}^{d}{a_{i}}^{k}} \right)^{\frac{1}{k}},~B_{d}Q_{d} = \left( {\sum_{i = 1}^{d}{b_{i}}^{k}} \right)^{\frac{1}{k}}
\end{aligned}
\label{eq11}
\end{equation}

Suppose that each dimension is independent and the values are uniformly distributed, so we can compute the expectation of $(A_d Q_d )^k/d$
\begin{equation}
\begin{aligned}
    \mathbf{E}\left( \left( \frac{A_{d}Q_{d}}{d^{\frac{1}{k}}} \right)^{k} \right) = \mathbf{E}\left( \left( \frac{B_{d}Q_{d}}{d^{\frac{1}{k}}} \right)^{k} \right)  
    = \mathbf{E}\left( {a_{i}}^{k} \right) = {\int_{0}^{1}{{a_{i}}^{k}da_{i}}} = \frac{1}{k + 1}
\end{aligned}
\label{eq12}
\end{equation}

Besides, we can calculate the variances of ${a_{i}}^{k}$ and ${b_{i}}^{k}$
\begin{equation}
\begin{aligned}
    var\left( {a_{i}}^{k} \right) &= var\left( {b_{i}}^{k} \right) = \mathbf{E}\left( {a_{i}}^{2k} \right) - \mathbf{E}^{2}\left( {a_{i}}^{k} \right)  
    \\&= {\int_{0}^{1}{{a_{i}}^{2k}da_{i}}} - \left( \frac{1}{k + 1} \right)^{2} = {\frac{1}{2k + 1}\left( \frac{k}{k + 1} \right)}^{2}
\end{aligned}
\label{eq13}
\end{equation}

Using Eq. \eqref{eq13}, the variance of $(A_d Q_d )^k/d$ can be obtained
\begin{equation}
\begin{aligned}
    \lim\limits_{d\rightarrow\infty}var\left( \left( \frac{A_{d}Q_{d}}{d^{\frac{1}{k}}} \right)^{k} \right) &= \lim\limits_{d\rightarrow\infty}var\left( \left( \frac{B_{d}Q_{d}}{d^{\frac{1}{k}}} \right)^{k} \right)  
    \\&= \lim\limits_{d\rightarrow\infty}\frac{1}{d^{2}}{\sum_{i = 1}^{d}{var\left( {a_{i}}^{k} \right)}} = 0
\end{aligned}
\label{eq14}
\end{equation}

Based on Lemma 1, Eq. \eqref{eq12} and Eq. \eqref{eq14}, we can deduce that
\begin{equation}
    \lim\limits_{d\rightarrow\infty}\frac{A_{d}Q_{d}}{d^{\frac{1}{k}}} = \lim\limits_{d\rightarrow\infty}\frac{B_{d}Q_{d}}{d^{\frac{1}{k}}} = \left( \frac{1}{k + 1} \right)^{\frac{1}{k}}
\label{eq15}
\end{equation}

Subsequently, we prove the Theorem 1 in two ways:

$i$) Since min and max are continuous functions, we conclude from Slutsky’s Theorem based on Eq. \eqref{eq15} that
\begin{equation}
    \lim\limits_{d\rightarrow\infty}\frac{D_{d,max}}{d^{\frac{1}{k}}} = \lim\limits_{d\rightarrow\infty}\frac{D_{d,min}}{d^{\frac{1}{k}}} = \left( \frac{1}{k + 1} \right)^{\frac{1}{k}}
\label{eq16}
\end{equation}

Therefore, we have
\begin{equation}
    \lim\limits_{d\rightarrow\infty}\frac{D_{d,max}}{D_{d,min}} = \lim\limits_{d\rightarrow\infty}\frac{D_{d,max}/d^{\frac{1}{k}}}{D_{d,min}/d^{\frac{1}{k}}} = 1
\label{eq17}
\end{equation}

Theorem 1 is hence proven.

$ii$) An alternative proof is also provided. By factoring polynomials, we can obtain
\begin{equation}
    \left| {A_{d}Q_{d} - B_{d}Q_{d}} \right| = \frac{\left| {{A_{d}Q_{d}}^{k} - {B_{d}Q_{d}}^{k}} \right|/d^{\frac{k - 1}{k}}}{\sum\limits_{i = 0}^{k - 1}{\left( \frac{A_{d}Q_{d}}{d^{\frac{1}{k}}} \right)^{k - r - 1} \cdot \left( \frac{B_{d}Q_{d}}{d^{\frac{1}{k}}} \right)^{r}}}
\label{eq18}
\end{equation}

Using Eq. \eqref{eq15}, we can get
\begin{equation}
    \underset{d\rightarrow\infty}\lim{\sum\limits_{r = 0}^{k - 1}{\left( \frac{A_{d}Q_{d}}{d^{1/k}} \right)^{k - r - 1} \cdot \left( \frac{B_{d}Q_{d}}{d^{1/k}} \right)^{r}}} = k\left( \frac{1}{k + 1} \right)^{\frac{k - 1}{k}}
\label{eq19}
\end{equation}

Based on the Central Limit Theorem, we have
\begin{equation}
    \left. {A_{d}Q_{d}}^{k} - {B_{d}Q_{d}}^{k} = {\sum\limits_{i = 1}^{d}\left( {{a_{i}}^{k} - {b_{i}}^{k}} \right)} \right.\sim\mathcal{N}\left( {0,~2d\sigma^{2}} \right)
\label{eq20}
\end{equation}

The variance can be obtained from Eq. \eqref{eq13}
\begin{equation}
    \sigma^{2} = var\left( {a_{i}}^{k} \right) = {\frac{1}{2k + 1}\left( \frac{k}{k + 1} \right)}^{2}
\label{eq21}
\end{equation}

The density of the normal distribution $\mathcal{N}\left( {0,~2d\sigma^{2}} \right)$ is
\begin{equation}
    f(x) = \frac{1}{2\sigma\sqrt{\pi d}}e^{- \frac{x^{2}}{4d\sigma^{2}}}
\label{eq22}
\end{equation}

The expectation of $\left| {A_{d}Q_{d}}^{k} - {B_{d}Q_{d}}^{k} \right|$ can be derived
\begin{equation}
\begin{aligned}
    \mathbf{E}( \left| {{A_{d}Q_{d}}^{k} - {B_{d}Q_{d}}^{k}} \right| ) &= {\int_{0}^{\infty}{2xf(x)dx}} 
    = {\int_{0}^{\infty}{\frac{x}{\sigma\sqrt{\pi d}}e^{- \frac{x^{2}}{4d\sigma^{2}}}dx}} \\&= 2\sqrt{\frac{d}{\pi\left( {2k + 1} \right)}}\left( \frac{k}{k + 1} \right) 
\end{aligned}
\label{eq23}
\end{equation}

Hence, we can compute the expectation of $\left| {A_{d}Q_{d} - B_{d}Q_{d}} \right|$

\begin{equation}
\begin{aligned}
    &\underset{d\rightarrow\infty}\lim\mathbf{E}\left( \left| {A_{d}Q_{d} - B_{d}Q_{d}} \right| \right) 
    \\&= \underset{d\rightarrow\infty}\lim\mathbf{E}\left( \frac{\left| {{A_{d}Q_{d}}^{k} - {B_{d}Q_{d}}^{k}} \right|/d^{\frac{k - 1}{k}}}{\sum\limits_{i = 0}^{k - 1}{\left( {A_{d}Q_{d}/d^{\frac{1}{k}}} \right)^{k - r - 1} \cdot \left( {B_{d}Q_{d}/d^{\frac{1}{k}}} \right)^{r}}} \right)
    \\&= \underset{d\rightarrow\infty}\lim\frac{2\sqrt{\frac{d}{\pi\left( {2k + 1} \right)}}\left( \frac{k}{k + 1} \right)/d^{\frac{k - 1}{k}}}{k\left( \frac{1}{k + 1} \right)^{\frac{k - 1}{k}}}
    = \underset{d\rightarrow\infty}\lim\frac{2}{\sqrt{\pi}}\sqrt{\frac{1}{2k + 1}}\frac{d^{\frac{1}{k} - \frac{1}{2}}}{\left( {k + 1} \right)^{\frac{1}{k}}}
\end{aligned}
\label{eq24}
\end{equation}

We take $\left| {A_{d}Q_{d} - B_{d}Q_{d}} \right|$ as a new variable, then its maximum is equal to $\left| {D_{d,max} - D_{d,min}} \right|$. Considering that there exists $n$ pair-wise distances $P_{d}^{i}Q_{d},i = 1,2,\ldots,n$, so $\left| {A_{d}Q_{d} - B_{d}Q_{d}} \right|$ contains $n(n-1)/2$ non-zero and non-repeating values at most and we can deduce that

\begin{equation}
\begin{aligned}
    \underset{d\rightarrow\infty}\lim\mathbf{E}\left( \left| {A_{d}Q_{d} - B_{d}Q_{d}} \right| \right) &\leq \underset{d\rightarrow\infty}\lim\left| {D_{d,max} - D_{d,min}} \right| \\&\leq \frac{n(n - 1)}{2}\underset{d\rightarrow\infty}\lim\mathbf{E}\left( \left| {A_{d}Q_{d} - B_{d}Q_{d}} \right| \right)
\end{aligned}
\label{eq25}
\end{equation}
\begin{equation}
\begin{aligned}
    \underset{d\rightarrow\infty}\lim\frac{2}{\sqrt{\pi}}\sqrt{\frac{1}{2k + 1}}\frac{d^{\frac{1}{k} - \frac{1}{2}}}{\left( {k + 1} \right)^{\frac{1}{k}}} &\leq \underset{d\rightarrow\infty}\lim\left| {D_{d,max} - D_{d,min}} \right| \\&\leq \underset{d\rightarrow\infty}\lim\frac{n(n - 1)}{\sqrt{\pi}}\sqrt{\frac{1}{2k + 1}}\frac{d^{\frac{1}{k} - \frac{1}{2}}}{\left( {k + 1} \right)^{\frac{1}{k}}}
\end{aligned}
\label{eq26}
\end{equation}

Besides, using Slutsky’s Theorem and Eq. \eqref{eq15}, we can obtain
\begin{equation}
    \underset{d\rightarrow\infty}\lim{D_{d,min}} = \underset{d\rightarrow\infty}\lim{A_{d}Q_{d}} = \left( \frac{d}{k + 1} \right)^{\frac{1}{k}}
\label{eq27}
\end{equation}

Then, we have
\begin{equation}
\begin{aligned}
    \underset{d\rightarrow\infty}\lim\frac{2}{\sqrt{\pi d(2k + 1)}} \leq \underset{d\rightarrow\infty}\lim\frac{\left| {D_{d,max} - D_{d,min}} \right|}{D_{d,min}} \leq \underset{d\rightarrow\infty}\lim\frac{n(n - 1)}{\sqrt{\pi d(2k + 1)}}
\end{aligned}
\label{eq28}
\end{equation}

For a given $k$ and $n$, we have
\begin{equation}
    \lim\limits_{d\rightarrow\infty}\frac{\left| {D_{d,max} - D_{d,min}} \right|}{D_{d,min}} = 0
\label{eq29}
\end{equation}

In fact, as described in Theorem 1, the value of each dimension of $P_d$ are not limited to being uniformly distributed in [0, 1], and the coordinates of $Q_d$ can also be any other point, rather than the origin only. Using the similar derivation above, the conclusion can be easily extended to more general scenarios.

\subsection{Chebyshev Distance}
Chebyshev distance is defined as the maximum absolute difference between the coordinates of the points across all dimensions, allowing it to emphasize the most significant variation between data points. It is also known as the $L_{\infty}$ distance measurement, since the $L_k$-norm Minkowski distance converges to the Chebyshev distance as k approaches to infinite. We can provide a brief proof as follow

\textbf{Proof:} Given two data points $A_{d} = \left( a_{1},a_{2},\ldots a_{d} \right) \in \mathbb{R}^{d}$ and $B_{d} = \left( b_{1},b_{2},\ldots b_{d} \right) \in \mathbb{R}^{d}$, we have
\begin{equation}
    {\max\limits_{i}\left| {a_{i} - b_{i}} \right|} \leq \lim\limits_{k\rightarrow\infty}\left( {\sum_{i = 1}^{d}\left| {a_{i} - b_{i}} \right|^{k}} \right)^{\frac{1}{k}}
     \leq \lim\limits_{k\rightarrow\infty}d^{\frac{1}{k}} \cdot {\max\limits_{i}\left| {a_{i} - b_{i}} \right|}
\label{eq30}
\end{equation}

For a given dimension $d$, we can conclude that

\begin{equation}
    \lim\limits_{k\rightarrow\infty}\left( {\sum_{i = 1}^{d}\left| {a_{i} - b_{i}} \right|^{k}} \right)^{\frac{1}{k}} = {\max\limits_{i}\left| {a_{i} - b_{i}} \right|}
\label{eq31}
\end{equation}

Similarly, Chebyshev distance also suffers from the distance concentration problem. Using NNS as the use case, we can formally depict this phenomenon as

\textbf{Theorem 2:} Given a data point $A_{d} = \left( a_{1},a_{2},\ldots a_{d} \right) \in \mathbb{R}^{d}$ and a query point $Q_{d} = (0,0,\ldots 0) \in \mathbb{R}^{d}$, if each dimension of $A_d$ is independent and $a_{i} \sim U\left( {s,~t} \right)$, for $0 \leq s \leq t$, the Chebyshev distance to the query point converges to $t$

\begin{equation}
    \lim\limits_{d\rightarrow\infty}{\max\limits_{i}a_{i}} = t
\label{eq32}
\end{equation}

\textbf{Proof:} We first compute the probability
\begin{equation}
\begin{aligned}
    P\left( {\max\limits_{i}a_{i}} \leq x \right) = {\prod\limits_{i = 1}^{d}{P\left( {a_{i} \leq x} \right)}} 
    = \left( {\int_{s}^{x}{\frac{1}{t - s}da_{i}}} \right)^{d} = \left( \frac{x - s}{t - s} \right)^{d}
\end{aligned}
\label{eq33}
\end{equation}

Let $z$ denote the Chebyshev distance, we can have the probability density
\begin{equation}
    f(z) = \frac{d(z - s)^{d - 1}}{(t - s)^{d}},~s \leq z \leq t
\label{eq34}
\end{equation}

Thus, we can compute the expectation
\begin{equation}
\begin{aligned}
    \mathbf{E}(z) &= {\int_{s}^{t}{z \cdot \frac{d\left( {z - s} \right)^{d - 1}}{\left( {t - s} \right)^{d}}dz}}
    \\&= \frac{d}{\left( {t - s} \right)^{d}}\left( {{\int_{s}^{t}{\left( {z - s} \right)^{d}dz}} + s{\int_{s}^{t}{\left( {z - s} \right)^{d - 1}dz}}} \right) 
    \\&= \frac{td + s}{d + 1}
\end{aligned}
\label{eq35}
\end{equation}

The limit of variance can also be obtained
\begin{equation}
\begin{aligned}
    \mathbf{E}\left( z^{2} \right) &= {\int_{s}^{t}{z^{2} \cdot \frac{d\left( {z - s} \right)^{d - 1}}{\left( {t - s} \right)^{d}}dz}}
    \\ &= \frac{d}{d + 2}\left( {t - s} \right)^{2} + \frac{2sd}{d + 2}\left( {t - s} \right) + s^{2}
\end{aligned}
\label{eq36}
\end{equation}
\begin{equation}
    \lim\limits_{d\rightarrow\infty}var(z) = \lim\limits_{d\rightarrow\infty}\mathbf{E}\left( z^{2} \right) - \lim\limits_{d\rightarrow\infty}\mathbf{E}^{2}(z) = 0
\label{eq37}
\end{equation}

In conclude, using Lemma 1, we can deduce that
\begin{equation}
   \lim\limits_{d\rightarrow\infty}{\max\limits_{i}a_{i}} = \lim\limits_{d\rightarrow\infty}\mathbf{E}(z) = \lim\limits_{d\rightarrow\infty}\frac{td + s}{d + 1} = t
\label{eq38}
\end{equation}

Under the assumption of independent uniform distribution, the limit of Chebyshev distance to the origin point is a constant that is equal to the upper bound of the uniform distribution. In fact, the query point $Q_d$ can be generalized to any position by translating the range of each dimension of the data point $A_d$.

\subsection{Cosine Distance}
Cosine distance has been extensively used on sparse and discrete domains for measuring the similarity of high-dimensional data. However, it could also be invalid such that any two vectors will be almost orthogonal with high probability. Inspired by \cite{ref5}, we formulate the distance concentration problem in cosine distance as the following theorem

\textbf{Theorem 3:} Given two data points $A_{d} = \left( a_{1},a_{2},\ldots a_{d} \right) \in \mathbb{R}^{d}$ and $B_{d} = \left( b_{1},b_{2},\ldots b_{d} \right) \in \mathbb{R}^{d}$, if each dimension is independent and $a_{i},b_{i} \sim U(s,~t)$, where $s<t$, then we have
\begin{equation}
\begin{aligned}
    \lim\limits_{d\rightarrow\infty}\cos\left\langle {A_{d},B_{d}} \right\rangle &= \lim\limits_{d\rightarrow\infty}\frac{\sum_{i = 1}^{d}{a_{i}b_{i}}}{\sqrt{\sum_{i = 1}^{d}{a_{i}}^{2}} \cdot \sqrt{\sum_{i = 1}^{d}{b_{i}}^{2}}}
    \\ &= \frac{3}{4}\left( 1 + \frac{st}{s^{2} + st + t^{2}} \right)
\end{aligned}
\label{eq39}
\end{equation}

\textbf{Proof:} We first compute the expectation and variance of random variable $a_i$
\begin{equation}
   \mathbf{E}\left( a_{i} \right) = {\int_{s}^{t}{\frac{a_{i}}{t - s}da_{i}}} = \frac{t + s}{2}
\label{eq40}
\end{equation}
\begin{equation}
   \mathbf{E}\left( {a_{i}}^{2} \right) = {\int_{s}^{t}{\frac{{a_{i}}^{2}}{t - s}da_{i}}} = \frac{s^{2} + st + t^{2}}{3}
\label{eq41}
\end{equation}
\begin{equation}
   var\left( a_{i} \right) = \mathbf{E}\left( {a_{i}}^{2} \right) - \mathbf{E}^{2}\left( a_{i} \right) = \frac{(t - s)^{2}}{12}
\label{eq42}
\end{equation}

For the sake of simplicity, we set
\begin{equation}
   u = \frac{1}{d}{\sum_{i = 1}^{d}{a_{i}b_{i}}},~~v = \frac{1}{d}{\sum_{i = 1}^{d}{a_{i}}^{2}}
\label{eq43}
\end{equation}

Using Eq. \eqref{eq40}, we can compute the expectation of $u$
\begin{equation}
   \mathbf{E}(u) = \mathbf{E}\left( {a_{i}b_{i}} \right) = \mathbf{E}^{2}\left( a_{i} \right) = \frac{(t + s)^{2}}{4}
\label{eq44}
\end{equation}

Using Eq. \eqref{eq40}-\eqref{eq44}, we obtain the limit of the variance of $u$
\begin{equation}
\begin{aligned}
    \lim\limits_{d\rightarrow\infty}var(u) &= \lim\limits_{d\rightarrow\infty}\frac{1}{d^{2}}{\sum_{i = 1}^{d}{var\left( {a_{i}b_{i}} \right)}}
    \\ &= \lim\limits_{d\rightarrow\infty}\frac{1}{d}\left( {{var}^{2}\left( a_{i} \right) + 2\mathbf{E}^{2}\left( a_{i} \right)var\left( a_{i} \right)} \right)
    \\ &= \lim\limits_{d\rightarrow\infty}\frac{\left( {t - s} \right)^{2}}{12d} \cdot \left( {\frac{\left( {t - s} \right)^{2}}{12} + \frac{\left( {t + s} \right)^{2}}{4}} \right) = 0
\end{aligned}
\label{eq45}
\end{equation}

Thus, based on Lemma 1, we can deduce that
\begin{equation}
   \lim\limits_{d\rightarrow\infty}u = \mathbf{E}(u) = \frac{(t + s)^{2}}{4}
\label{eq46}
\end{equation}

Similarly, we have
\begin{equation}
   \lim\limits_{d\rightarrow\infty}v = \mathbf{E}(v) = \mathbf{E}\left( {a_{i}}^{2} \right) = \frac{s^{2} + st + t^{2}}{3}
\label{eq47}
\end{equation}

Since $a_i$ and $b_i$ are IID, we can conclude that
\begin{equation}
\begin{aligned}
    &\lim\limits_{d\rightarrow\infty}\cos\left\langle {A_{d},B_{d}} \right\rangle = \lim\limits_{d\rightarrow\infty}\frac{\sum_{i = 1}^{d}{a_{i}b_{i}}}{\sqrt{\sum_{i = 1}^{d}{a_{i}}^{2}} \cdot \sqrt{\sum_{i = 1}^{d}{b_{i}}^{2}}}
    \\&= \lim\limits_{d\rightarrow\infty}\frac{\frac{1}{d}{\sum_{i = 1}^{d}{a_{i}b_{i}}}}{\sqrt{\frac{1}{d}{\sum_{i = 1}^{d}{a_{i}}^{2}}} \cdot \sqrt{\frac{1}{d}{\sum_{i = 1}^{d}{b_{i}}^{2}}}} = \frac{\lim\limits_{d\rightarrow\infty}u}{\sqrt{\lim\limits_{d\rightarrow\infty}v} \cdot \sqrt{\lim\limits_{d\rightarrow\infty}v}}
    \\&= \frac{3}{4}\left( {1 + \frac{st}{s^{2} + st + t^{2}}} \right)
\end{aligned}
\label{eq48}
\end{equation}

As the above derivation, the cosine distance converges to a constant as the $d$ increases. In practice, we can consider the values of each dimension are uniformly distributed in the range of $[-c,~c]$ for symmetry, then the limit of cosine distance is zero in this case using Eq. \eqref{eq48}. It means that any two vectors tend to become orthogonal to each other in high-dimensional space.

\subsection{Empirical Analysis}
\begin{figure}[h]
\includegraphics[width=0.9\linewidth]{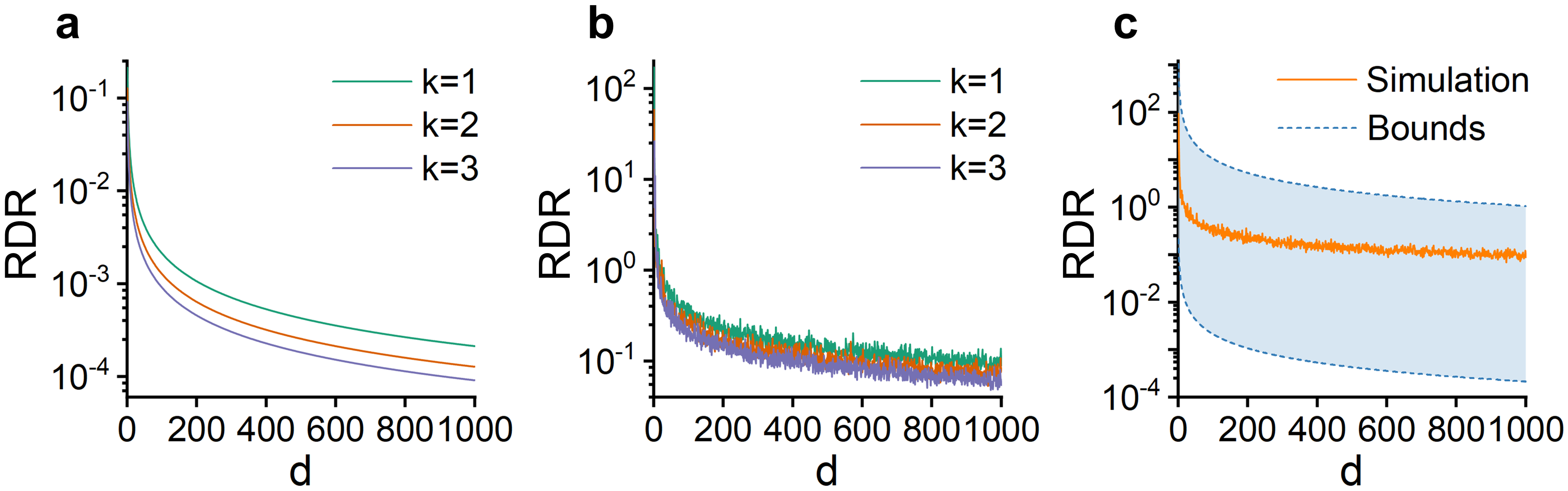}
\centering
\caption{Relative distance ratio of the Minkowski distance with 100 points and different norms. (a) Trends of the lower bound in Eq. \eqref{eq28} and (b) the simulation results under different dimensions. (c) The bounds in Eq. \eqref{eq28} and simulated results under $k=1$.
}
\label{fig4}

\end{figure}

\begin{figure}[h]
\includegraphics[width=0.9\linewidth]{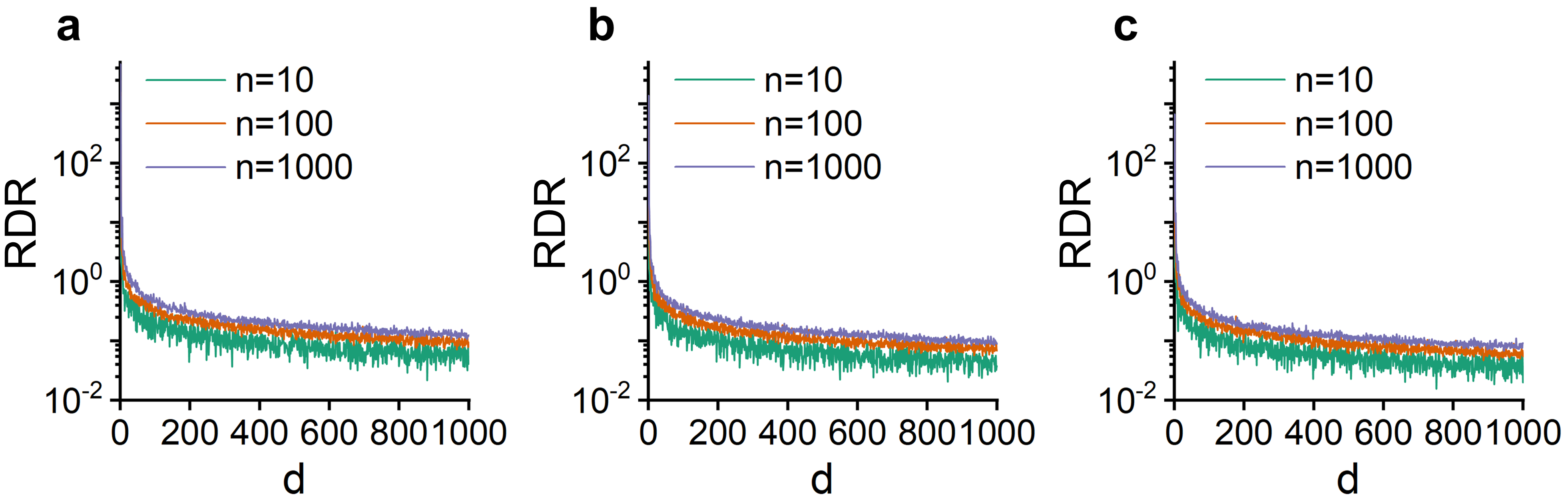}
\centering
\caption{ Relative distance ratio of the Minkowski distance using different numbers of points with a norm of (a) $k=1$, (b) $k=2$, and (c) $k=3$, respectively. 
}
\label{fig5}
\end{figure}

\textbf{Minkowski Distance.} To demonstrate Theorem 1, we further performed an simulation experiment by randomly generating 10, 100, 1000 data points of different dimensions ($d$) and ensuring that the value of each dimension is uniformly distributed in [0, 1]. We used 100 points to investigate the concentration of the Minkowski distance with different norms ($k$). The trends of the lower bound in Eq. \eqref{eq28} and simulation results are presented in Fig. \ref{fig4}\textcolor{blue}{a} and \textcolor{blue}{b}, respectively. It can be found that there exists concentration phenomenon under different norms of the Minkowski distance, and a higher norm causes more significant distance concentration. Fig. \ref{fig4}\textcolor{blue}{c} illustrates that the simulated RDE lies between the lower and upper bounds in Eq. \eqref{eq28}, which demonstrates the validity of the bound estimation. Meanwhile, we explored the distance concentration under different number of data points in Fig. \ref{fig5}. The more data points, the weaker distance concentration effect. Although utilizing a lower norm and more data samples can alleviate the distance concentration, the effect is limited.


\begin{figure}[h]
\includegraphics[width=0.9\linewidth]{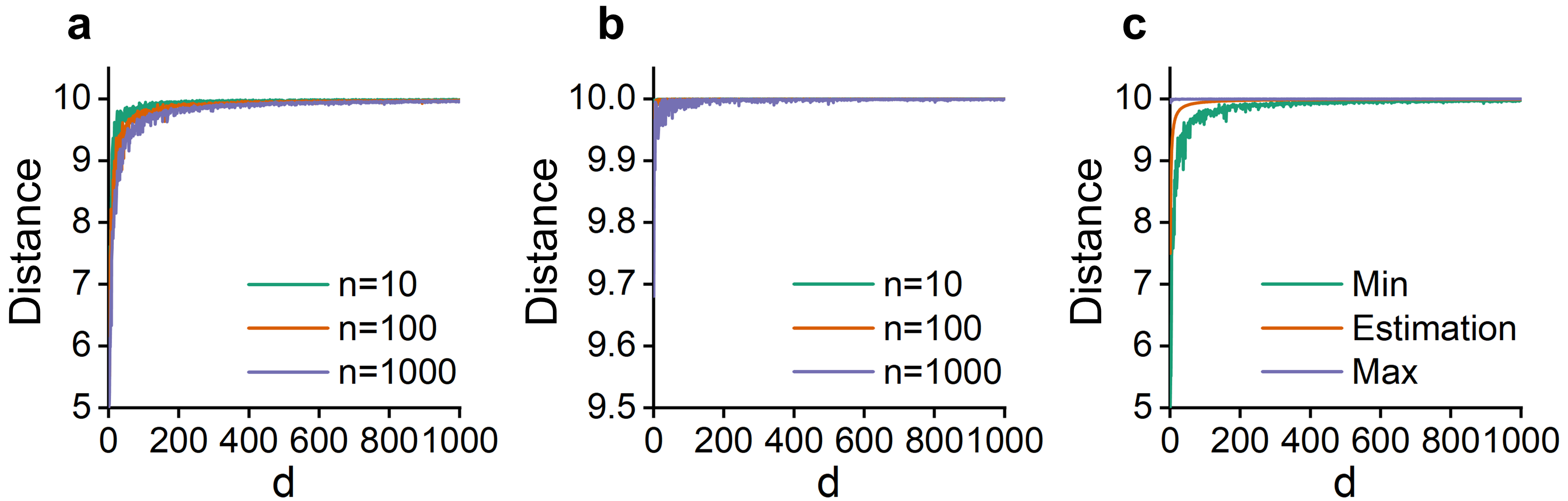}
\centering
\caption{Distance concentration in the Chebyshev distance. (a) The minimum and (b) maximum distances as the dimension $d$ grows under different data sizes. (c) Trends of the estimated, minimum and maximum distances by varying the dimension.}
\label{fig6}
\end{figure}

\textbf{Chebyshev Distance.} We generated the simulated data with the value of each dimension being uniformly distributed in [5, 10] ($s=5, ~t=10$). As presented in Fig. \ref{fig6}\textcolor{blue}{a} and \textcolor{blue}{b}, the minimum distance is close to 5 (the lower bound) when $d$ is low, and both the minimum and maximum distances approach 10 (the upper bound) with the growth of dimensionality. This pattern coincides with Eq. \eqref{eq38}. The larger sample size, the slower convergence speed. We also compared the simulation results with the estimation results using $(td+s)/(d+1)$ 
in Eq. \eqref{eq38}. Fig. \ref{fig6}\textcolor{blue}{c} demonstrates the consistency between them.

\begin{figure}[h]
\includegraphics[width=0.9\linewidth]{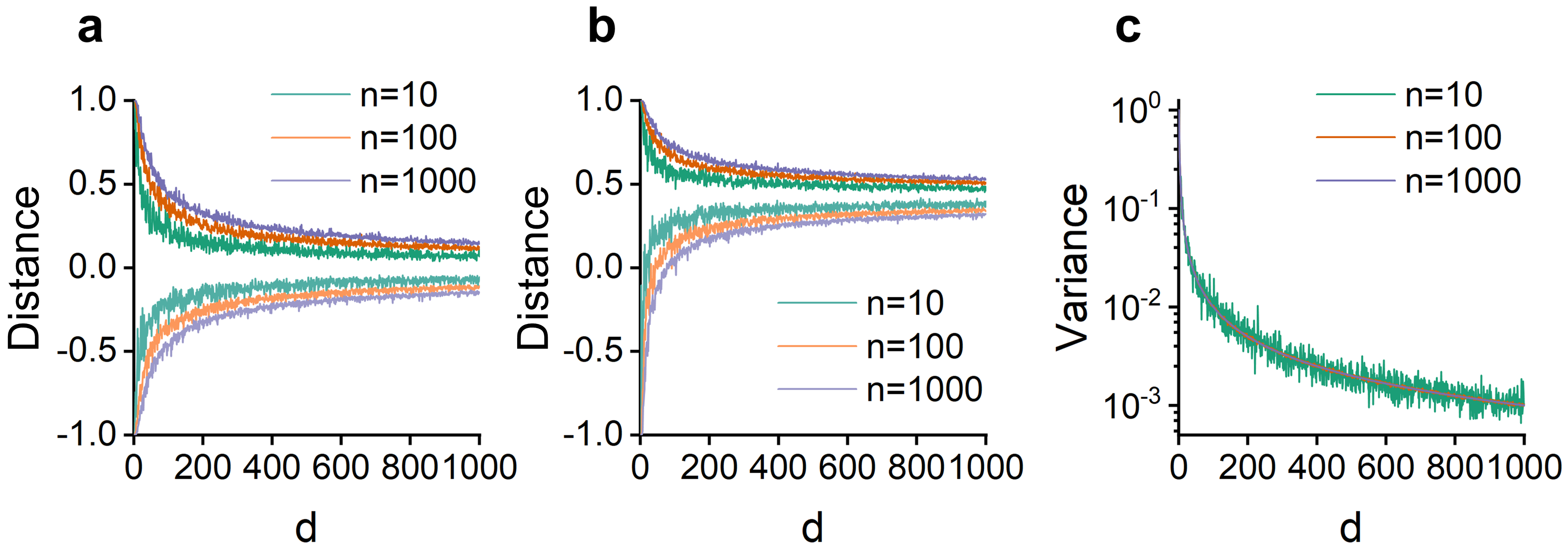}
\centering
\caption{Distance concentration in the cosine distance. Trends of the simulated cosine distance with (a) $s=-1,t=1$ and (b) $s=-1,t=3$, where the top three and bottom three lines denote the maximum and minimum distances respectively.(c) Trends of the variances by varying the dimension.}
\label{fig7}
\end{figure}

\textbf{Cosine Distance.} We computed the simulated cosine distances by setting the value of each dimension range from [-1, 1] (Fig. \ref{fig7}\textcolor{blue}{a}) and [-1, 3] (Fig. \ref{fig7}\textcolor{blue}{b}). The minimum and maximum cosine distances converge to the same value as the dimension grows and the variance approach zero in Fig. \ref{fig7}\textcolor{blue}{c}, which presents the consistent law with Eq. \eqref{eq48}. Like other distance metrics, an exponential increase in data size has a limited effect for slowing down the distance concentration.

\section{Manifold Effect}
\label{sec4}
High-dimensional data often exhibits a non-linear manifold structure, which refers to the property where the intrinsic dimension of data is lower than the feature dimension. Manifold effect exists in high-dimensional space and is more remarkable when the number of samples is much smaller than the feature dimension. It is also called the high dimension low sample size (HDLSS) problem ($d \gg n$), and the asymptotic behavior ($\left. d\rightarrow\infty \right.$) has been studied for exploring the PCA consistency \cite{ref27}, \cite{ref28}, \cite{ref29}, \cite{ref30}. The eigenvalues also called variances in PCA can reflect the magnitude of information in each dimension of the data, thereby revealing the existence of manifold effect. In this section, we investigate the asymptotic behavior of the cumulative contribution ratio (CCR) of eigenvalue in PCA and provide a new perspective on the manifold effect. 

\subsection{Theoretical Analysis}
\textbf{Definition:} Let $\mathbf{X} = \left( \mathbf{x}_{1},\mathbf{x}_{2},\ldots\mathbf{x}_{n} \right) \in \mathbb{R}^{d \times n}$ be the data matrix that holds $n$ random and independent samples $\mathbf{x}_{i} = \left( x_{1i},x_{2i},\ldots x_{di} \right)^{T}$. If $\mathbf{\mu} = \left( \mu_{1},\mu_{2},\ldots\mu_{d} \right)^{T}$ denotes the mean of each row of $\mathbf{X}$, where $\mu_{i} = {\sum_{j = 1}^{n}x_{ij}}/n$, the population covariance matrix is defined as
\begin{equation}
\begin{aligned}
    \mathbf{C} = \begin{pmatrix}
    \begin{matrix}
    c_{11} & c_{12} \\
    c_{21} & c_{22}
    \end{matrix} & \begin{matrix}
    \ldots & c_{1d} \\
    \ldots & c_{2d}
    \end{matrix} \\
    \begin{matrix}
     \vdots & \vdots \\
    c_{d1} & c_{d2}
    \end{matrix} & \begin{matrix}
    ~ & \vdots \\
    \ldots & c_{dd}
    \end{matrix}
    \end{pmatrix}
    ,~
    c_{ij} = \frac{1}{n}{\sum_{l = 1}^{n}{\left( {x_{il} - \mu_{i}} \right)\left( {x_{jl} - \mu_{j}} \right)}}
\end{aligned}
\label{eq49}
\end{equation}
\textbf{Theorem 4:} Suppose that the elements in $\mathbf{X}$ are independent and identically distributed in a finite range with the fixed expectation $\mathbf{E}(x)$ and variance $var(x)$, and the eigenvalues of $\mathbf{C}$ are arranged as $0 \leq \lambda_{1} \leq \lambda_{2} \leq \ldots \leq \lambda_{d}$, if $d \gg n$, then we have that the CCR of the first $d-n$ smallest eigenvalues converges to zero, which can be stated as
\begin{equation}
\underset{d\rightarrow\infty}\lim{CCR} = \underset{d\rightarrow\infty}\lim\frac{\sum_{i = 1}^{d - n}\lambda_{i}}{\sum_{i = 1}^{d}\lambda_{i}} = 0
\label{eq50}
\end{equation}
\textbf{Proof:} We compute the limits of $\sum_{i = 1}^{d}\lambda_{i}$ (\textit{Part I}) and $\sum_{i = 1}^{d - n}\lambda_{i}$ (\textit{Part II}) separately as follows

\textit{Part I:} The sum of all eigenvalues is equal to the sum of the diagonal elements of $\mathbf{C}$
\begin{equation}
\begin{aligned}
{\sum\limits_{i = 1}^{d}\lambda_{i}} &= {\sum\limits_{i = 1}^{d}c_{ii}} = \frac{1}{n}{\sum\limits_{i = 1}^{d}{\sum\limits_{j = 1}^{n}\left( {x_{ij} - \mu_{i}} \right)^{2}}}
= \frac{1}{n^{2}}{\sum\limits_{i = 1}^{d}\left( {{\sum\limits_{j = 1}^{n}{x_{ij}}^{2}} - n{\mu_{i}}^{2}} \right)} \\ &= \frac{1}{n^{2}}{\sum\limits_{i = 1}^{d}{\sum\limits_{j \neq l}^{n}\left( {x_{ij} - x_{il}} \right)^{2}}}
= \frac{1}{n^{2}}{\sum\limits_{i = 1}^{d}{\sum\limits_{j \neq l}^{n}\left( {{x_{ij}}^{2} + {x_{il}}^{2} - 2x_{ij}x_{il}} \right)}}
\end{aligned}
\label{eq51}
\end{equation}

Hence, we can obtain the limit of the expectation 
\begin{equation}
\begin{aligned}
\underset{d\rightarrow\infty}\lim\mathbf{E}\left( {\frac{1}{d}{\sum_{i = 1}^{d}\lambda_{i}}} \right)
& = \underset{d\rightarrow\infty}\lim\frac{1}{dn^{2}}\mathbf{E}\left( {\sum_{i = 1}^{d}{\sum_{j \neq l}^{n}\left( {{x_{ij}}^{2} + {x_{il}}^{2} - 2x_{ij}x_{il}} \right)}} \right)
\\ &= \underset{d\rightarrow\infty}\lim\frac{n - 1}{n}\left( {\mathbf{E}\left( x^{2} \right) - \mathbf{E}^{2}(x)} \right) = \frac{n - 1}{n}var(x)
\end{aligned}
\label{eq52}
\end{equation}

We can compute the limit of the variance
\begin{equation}
\begin{aligned}
0 &\leq \underset{d\rightarrow\infty}\lim{var}\left( {\frac{1}{d}{\sum_{i = 1}^{d}\lambda_{i}}} \right)
\\ &= \underset{d\rightarrow\infty}\lim\frac{n - 1}{dn}\left( {var\left( x^{2} \right) - {var}^{2}(x) - 2\mathbf{E}^{2}(x)var(x)} \right)
\\ &\leq \underset{d\rightarrow\infty}\lim\frac{n - 1}{dn}\left( {\mathbf{E}\left( x^{4} \right)} \right) = 0
\end{aligned}
\label{eq53}
\end{equation}

Based on Lemma 1, Eq. \eqref{eq52} and \eqref{eq53}, we conclude that
\begin{equation}
\underset{d\rightarrow\infty}\lim\frac{1}{d}{\sum_{i = 1}^{d}\lambda_{i}} = \frac{n - 1}{n}var(x)
\label{eq54}
\end{equation}

\textit{Part II:} Given any eigenvalue $\lambda$ of $\mathbf{C}$ and its corresponding eigenvector $\mathbf{v} = \left( {v_{1},v_{2},\ldots v_{d}} \right)^{T}$, $\mathbf{v}^{T}\mathbf{v} = 1$, we have
\begin{equation}
\begin{aligned}
\lambda &= \mathbf{v}^{T}\mathbf{C}\mathbf{v} = \frac{1}{n}{\sum_{i = 1}^{d}{\sum_{j = 1}^{d}{v_{i}v_{j}{\sum_{l = 1}^{n}{\left( {x_{il} - \mu_{i}} \right)\left( {x_{jl} - \mu_{j}} \right)}}}}}
\\&= \frac{1}{n}{\sum_{l = 1}^{n}{\sum_{i = 1}^{d}{v_{i}\left( {x_{il} - \mu_{i}} \right)v_{j}{\sum_{j = 1}^{d}\left( {x_{jl} - \mu_{j}} \right)}}}}
\\&= \frac{1}{n}{\sum_{l = 1}^{n}{\sum_{i = 1}^{d}{v_{i}\left( {x_{il} - \mu_{i}} \right)\mathbf{v}^{T}\left( {\mathbf{x}_{l} - \mathbf{\mu}} \right)}}}
= \frac{1}{n}{\sum_{l = 1}^{n}\left( {\mathbf{v}^{T}\left( {\mathbf{x}_{l} - \mathbf{\mu}} \right)} \right)^{2}}
\end{aligned}
\label{eq55}
\end{equation}

We assume $\tilde{\mathbf{X}} = \left( \mathbf{x}_{1} - \mathbf{\mu},\mathbf{x}_{2} - \mathbf{\mu},\ldots\mathbf{x}_{n} - \mathbf{\mu} \right) \in \mathbb{R}^{d \times n}$ is the mean-centered data matrix, and consider the homogeneous system of linear equations $\mathbf{v}^{T}\tilde{\mathbf{X}} = 0$
\begin{equation}
\begin{aligned}
\left\{ \begin{matrix}
{\mathbf{v}^{T}\left( {\mathbf{x}_{1} - \mathbf{\mu}} \right) = 0} \\
{\mathbf{v}^{T}\left( {\mathbf{x}_{2} - \mathbf{\mu}} \right) = 0} \\
 \vdots \\
{\mathbf{v}^{T}\left( {\mathbf{x}_{n} - \mathbf{\mu}} \right) = 0}
\end{matrix} \right.
\end{aligned}
\label{eq56}
\end{equation}

For $rank\left( \tilde{\mathbf{X}} \right) \leq n \ll d$, that means this system has $d-n$ basic solutions at least. By Gram-Schmidt orthogonalization, $d-n$ orthonormal vectors $\mathbf{v}_{1},\mathbf{v}_{2},\ldots\mathbf{v}_{d - n}$ can be solved from the linearly independent solutions. These vectors are the eigenvectors of the covariance matrix $\mathbf{C}$, and corresponds the $d-n$ smallest eigenvalues, thus we have
\begin{equation}
\lambda_{1} = \lambda_{2} = \ldots = \lambda_{d - n} = 0
\label{eq422}
\end{equation}

Combining Eq. \eqref{eq54} and \eqref{eq422}, we conclude that
\begin{equation}
\underset{d\rightarrow\infty}\lim\frac{\sum_{i = 1}^{d - n}\lambda_{i}}{\sum_{i = 1}^{d}\lambda_{i}} = \frac{\frac{1}{d}{\sum_{i = 1}^{d - n}\lambda_{i}}}{\frac{1}{d}{\sum_{i = 1}^{d}\lambda_{i}}} = 0
\label{eq57}
\end{equation}

Eq. \eqref{eq57} indicates that the $d$-dimensional data can be 100\% explained using only $n$ principal components (PC). It means that the intrinsic dimension of data is lower than $n$.

Manifolds exist in data not only when $d>n$. In fact, $rank( \tilde{\mathbf{X}})$ is always smaller than $d$ in real-world data even when $d \leq n$. This means that the system of equations in Eq. \eqref{eq56} has more than one solution, and there exists more than one zero eigenvalue of $\mathbf{C}$ accordingly, indicating that the data is still a manifold. A weaker conclusion is that the contribution of the first smallest eigenvalue tends to decrease monotonously as $d$ increases, which can be depicted as
\begin{equation}
\frac{\lambda_{1}\left( \mathbf{C}^{(d)} \right)}{\sum_{i = 1}^{d}{\lambda_{i}\left( \mathbf{C}^{(d)} \right)}} \leq \frac{\lambda_{1}\left( \mathbf{C}^{({d - 1})} \right)}{\sum_{i = 1}^{d - 1}{\lambda_{i}\left( \mathbf{C}^{({d - 1})} \right)}} \leq \ldots \leq \frac{\lambda_{1}\left( \mathbf{C}^{(1)} \right)}{\sum_{i = 1}^{1}{\lambda_{i}\left( \mathbf{C}^{(1)} \right)}}
\label{eq58}
\end{equation}
where $\mathbf{C}^{(d)}$ represents the covariance matrix of a $d$-dimensional data. Using Cauchy Interlace Theorem, Eq. \eqref{eq58} can be easily proved. It illustrates that the manifold effect becomes more pronounced as the dimensionality increases.

\subsection{Empirical Analysis}
\begin{figure}[h]
\includegraphics[width=0.7\linewidth]{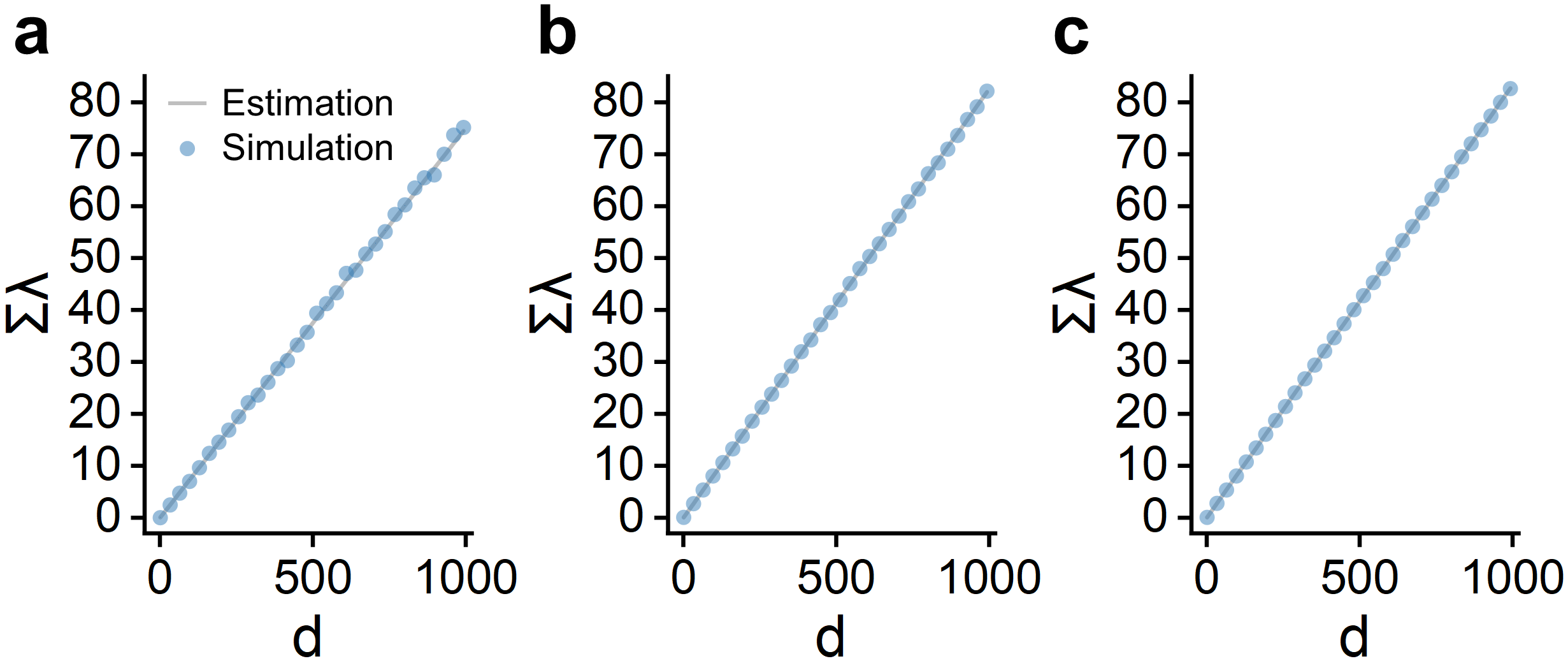}
\centering
\caption{Sum of all eigenvalues of the covariance matrix as the dimension grows
using different numbers of data points, (a) $n=100$, (b) $n=500$, (c) $n=1000$, respectively.}
\label{fig8}
\end{figure}

To demonstrate the conclusion in Eq. \eqref{eq54}, we conducted a simulation experiment by randomly generating 100, 500, and 1000 points and dimensions from 1 to 1000. The value of each feature is uniformly distributed in [0, 1] with the variance of 1/12. Fig. \ref{fig8} presents that the estimated results in Eq. \eqref{eq54} coincide with the simulated results as the $d$ grows, and the consistency becomes stronger with the increase of the data size. 

\begin{figure}[h]
\includegraphics[width=0.7\linewidth]{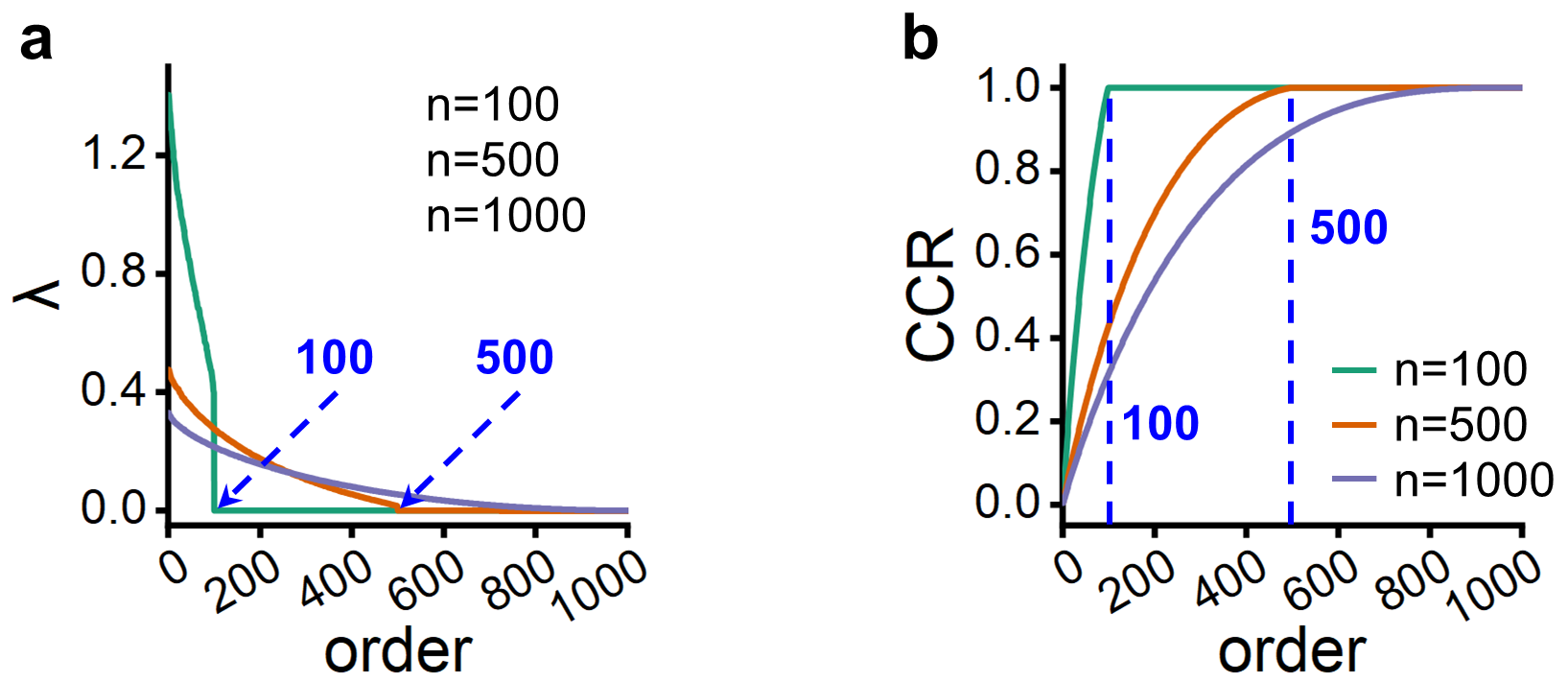}
\centering
\caption{The eigenvalues in descending order (b) and corresponding CCRs (c) in 1000-D simulated datasets with different sample sizes.}
\label{fig9}
\end{figure}

We verified Eq. \eqref{eq422} on the simulated datasets with different sizes. The descending-ordered eigenvalues of the covariance matrix and the corresponding CCRs are shown in Fig. \ref{fig9}\textcolor{blue}{b} and \textcolor{blue}{c} respectively, where the $d-n$ smallest eigenvalues are zero and CCR of the $n$ largest eigenvalues is equal to one, exhibiting the same law with Theorem 4.

\begin{figure}[h]
\includegraphics[width=1\linewidth]{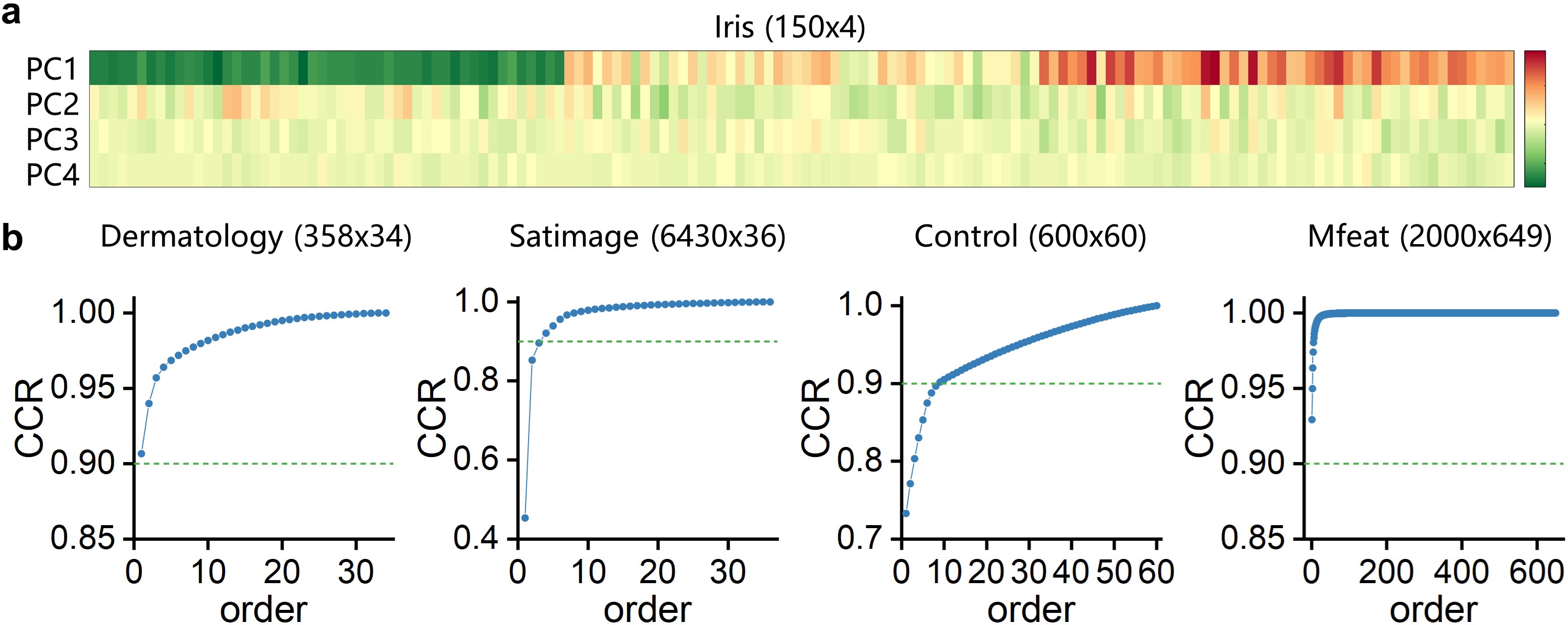}
\centering
\caption{PCA results on five UCI datasets. (a) A heatmap of Iris dataset after PCA rotation. (b) CCR of the eigenvalues in descending
order on Dermatology, Satimage, Control, and Mfeat, where the green line denotes the CCR=0.9.}
\label{fig10}
\end{figure}

Moreover, we performed PCA on five real-world UCI datasets, including Iris, Dermatology, Satimage, Control, and Mfeat \cite{ref31}. The first PC captures most of the discriminative information of Iris, but the other three PCs present a weak signal in Fig. \ref{fig10}\textcolor{blue}{a}. Meanwhile,
the CCR trends of the four datasets in Fig. \ref{fig10}\textcolor{blue}{b} implies that the above data can be explained 90\% using only less ten PCs. It indicates that the manifold effect is common in practical scenarios, and redundant features exist in the real-world data.

\section{Conclusion}
\label{sec5}
Curse of dimensionality impedes the effectiveness to train machine learning models and identify clustering patterns from the high-dimensional data. This paper aims to excavate the underlying causes of the curse of dimensionality, especially for distance concentration and manifold effect. Through theoretical and empirical analyses, we revealed the existences and patterns of distance concentration and manifold effect. As the dimension increases, distance measurement would be invalid and data exhibits non-linear manifolds with some redundant features. Although expansion of the data size slows down the convergence speed of distance concentration, the available amount of data in practical applications is limited. To mitigate the curse of dimensionality, dimension reduction techniques, such as PCA or the cutting-edge manifold learning techniques like t-SNE and UMAP, can be employed to reduce the number of dimensions while preserving the most important information. Meanwhile, careful feature selection, regularization techniques, and domain knowledge can help partially address the challenges associated with high-dimensional data.

\section*{Acknowledgements}
This work was supported by National Natural Science Foundation of China (No. 42090010, No. 41971349, No. 41930107) and the Fundamental Research Funds for the Central Universities, China (No. 2042022dx0001).

\section*{Data Availability}
The real-world datasets used in this study are publicly available: 

\begin{itemize}
  \item Iris (http://archive.ics.uci.edu/ml/datasets/Iris)
  \item Dermatology (http://archive.ics.uci.edu/dataset/33/dermatology) 
  \item Satimage (http://archive.ics.uci.edu/dataset/146/statlog+landsat+satellite)
  \item Control (http://archive.ics.uci.edu/ml/datasets/Synthetic+Control+Chart+Time+Series)
  \item Mfeat (https://archive.ics.uci.edu/dataset/72/multiple+features)
\end{itemize}


\end{document}